\documentclass{article}

\usepackage[final]{nips_2018}       

\usepackage[utf8]{inputenc} 
\usepackage[T1]{fontenc}    
\usepackage{hyperref}       
\usepackage{url}            
\usepackage{booktabs}       
\usepackage{amsfonts}       
\usepackage{nicefrac}       
\usepackage{microtype}      

\usepackage{epsfig}
\usepackage{graphics}
\usepackage{multirow}

\title{Visual Entailment Task for Visually-Grounded Language Learning}

\author{
  Ning Xie\thanks{Work performed as a NEC Labs intern}\\
  Wright State University\\
  Dayton, OH 45435\\
  \texttt{xie.25@wright.edu}\\
  \And
  Farley Lai \\
  NEC Laboratories America \\
  Princeton, NJ 08540 \\
  \texttt{farleylai@nec-labs.com} \\
  \And
  Derek Doran \\
  Wright State University \\
  Dayton, OH 45435\\
  \texttt{derek.doran@wright.edu} \\
\And
  Asim Kadav \\
  NEC Laboratories America \\
  Princeton, NJ 08540 \\
  \texttt{asim@nec-labs.com} \\
}

\begin{document}

\maketitle

\begin{abstract}

  We introduce a new inference task - \textbf{Visual Entailment (VE)} - which differs from traditional Textual Entailment (TE) tasks whereby a premise is defined by an image, rather than a natural language sentence as in TE tasks.
  A novel dataset \emph{SNLI-VE} (publicly available at \url{https://github.com/necla-ml/SNLI-VE}) is proposed for VE tasks based on the Stanford Natural Language Inference corpus and Flickr30k. 
  We introduce a differentiable architecture called the Explainable Visual Entailment model (EVE) to tackle the VE problem.
  EVE and several other state-of-the-art visual question answering (VQA) based models are evaluated on the SNLI-VE dataset, facilitating grounded language understanding and providing insights on how modern VQA based models perform.

\end{abstract}

\section{Introduction}

Multimodal inference, reasoning, and fact entailment across image data and text have the potential to solve problems where the veracity of a text statement is drawn from visual facts.
Representative applications involve the fake news detection and court cross-examination.
The former aims to detect contradictions between the text news and visual facts such as an image or video clip in order to reduce the influence of misleading news.
The latter intends to validate the testimony in case of any contradictions to visual evidence for a fair judgment.

Recent progress in visual reasoning using datasets such as the Visual Question Answering (VQA) dataset~\citep{antol2015vqa} and CLEVR~\citep{johnson2017clevr} has been encouraging. 
However, the high accuracy in these datasets is often because of the bias in these datasets.
For the VQA dataset, there is a question-conditioned bias~\citep{goyal2017making} where questions may hint at the answers such that the correct answer may be inferred without even considering the visual information.
The following version of the VQA dataset~\citep{goyal2017making} reduces the bias by pairing questions with similar images that lead to different answers.
Even so, the sentence structures in the VQA dataset remain simple and the yes/no questions are insufficient for training entailment tasks that include the neutral case.
CLEVR on the other hand is designed for fine-grained reasoning but its synthetic nature introduces the uniformity in image and text structures, resulting in very high accuracy models~\citep{hudson2018compositional} that may not generalize well to real world settings.
Hence, we need a more challenging inference task that requires learning grounded representations from cross-modal (image, text) pairs, where the same image is used for multiple natural language sentences, each of which may correspond to different answers.
Derived from this motivation, we propose a new \emph{Visual Entailment} (VE) task in this paper.

\begin{figure}[h]
  \centering
  \includegraphics[width=\linewidth]{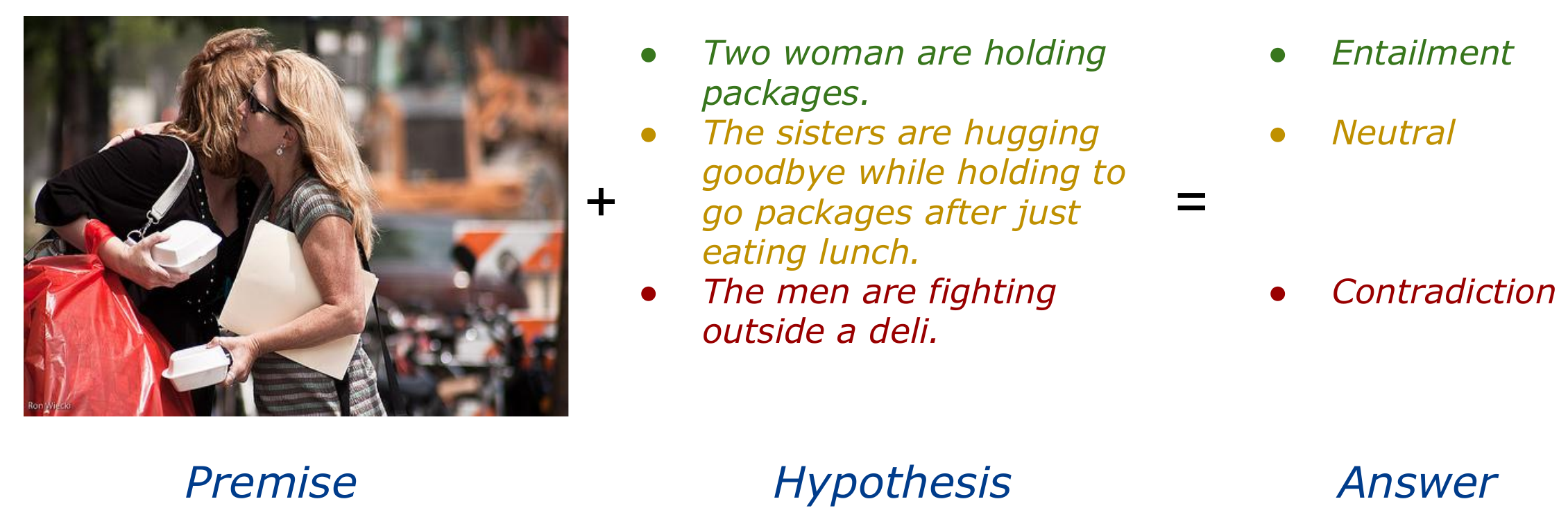}
  \caption{A VE example showing an image pairing with different hypotheses leads to different labels.}
  \label{fig:SNLI-VE}
\end{figure}

Prior to VE, the Textual Entailment (TE) task has been extensively studied in the natural language processing (NLP) community as part of natural language inference (NLI). 
In the TE task, given a text premise $P$ and a text hypothesis $H$, the goal is to determine if $P$ implies $H$. 
A TE model outputs a label out of the three classes: \emph{entailment}, {\em neutral} or \emph{contradiction} based on the relation conveyed by the $(P, H)$ text pair.
\emph{Entailment} holds if there is enough evidence in $P$ to conclude that $H$ is true.
\emph{Contradiction} is concluded wherever $H$ contradicts $P$.
Otherwise, the relation is \emph{neutral}, indicating the evidence in $P$ is insufficient to draw a conclusion from $H$.
We extend TE to the visual domain by replacing each text premise with a corresponding real world image.
Figure~\ref{fig:SNLI-VE} illustrates a VE example where given an image premise,
the three text hypotheses lead to three different class labels.

In contrast to existing yes/no VQA problems, our VE task is more challenging for requiring the model to deduce the \emph{neutral} case due to insufficient information.
To the best of our knowledge, there is no well-annotated dataset for VE. 
We then build a new dataset, SNLI-VE, by replacing the premises in the Stanford Natural Language Inference corpus (SNLI)~\citep{bowman2015large}, a TE dataset, with the corresponding images in Flickr30k~\citep{young2014image}, an image captioning dataset.
This adaption is possible since the premises in SNLI are from the Flickr30k image captions which are entailed by the corresponding images automatically.
By transitivity of entailment, those hypotheses entailed by the text premises are also entailed by the original caption images.
There is a chance that neutral and contradiction relations may change because the images may include other entities that unexpectedly rewrite neutral and contradiction conclusions.
Recently work \citep{vu2018grounded} combining both images and captions as premises validates that the effects of conclusion changes happen to be tolerable.

\paragraph{Related work.} The most relevant task to VE is VQA~\citep{antol2015vqa, goyal2017making, zhu2016visual7w, ren2015exploring, johnson2017inferring, hudson2018compositional, anderson2018bottom, fukui2016multimodal}, which is a representative multimodal task in machine learning that involves both images and text.
State-of-the-art VQA models commonly apply the attention mechanism~\citep{kim2018bilinear, anderson2018bottom, hudson2018compositional} to relate image regions with specific text features.
Our developed model tackles the VE task by further employing \emph{self-attention}~\citep{vaswani2017attention} to find the inner relationships in both image and text feature spaces as well as \emph{text-image attention} to ground relevant image regions.

\section{The EVE Architecture}
\label{sec:eve}

\begin{figure}[h]
  \centering
  \includegraphics[width=\linewidth]{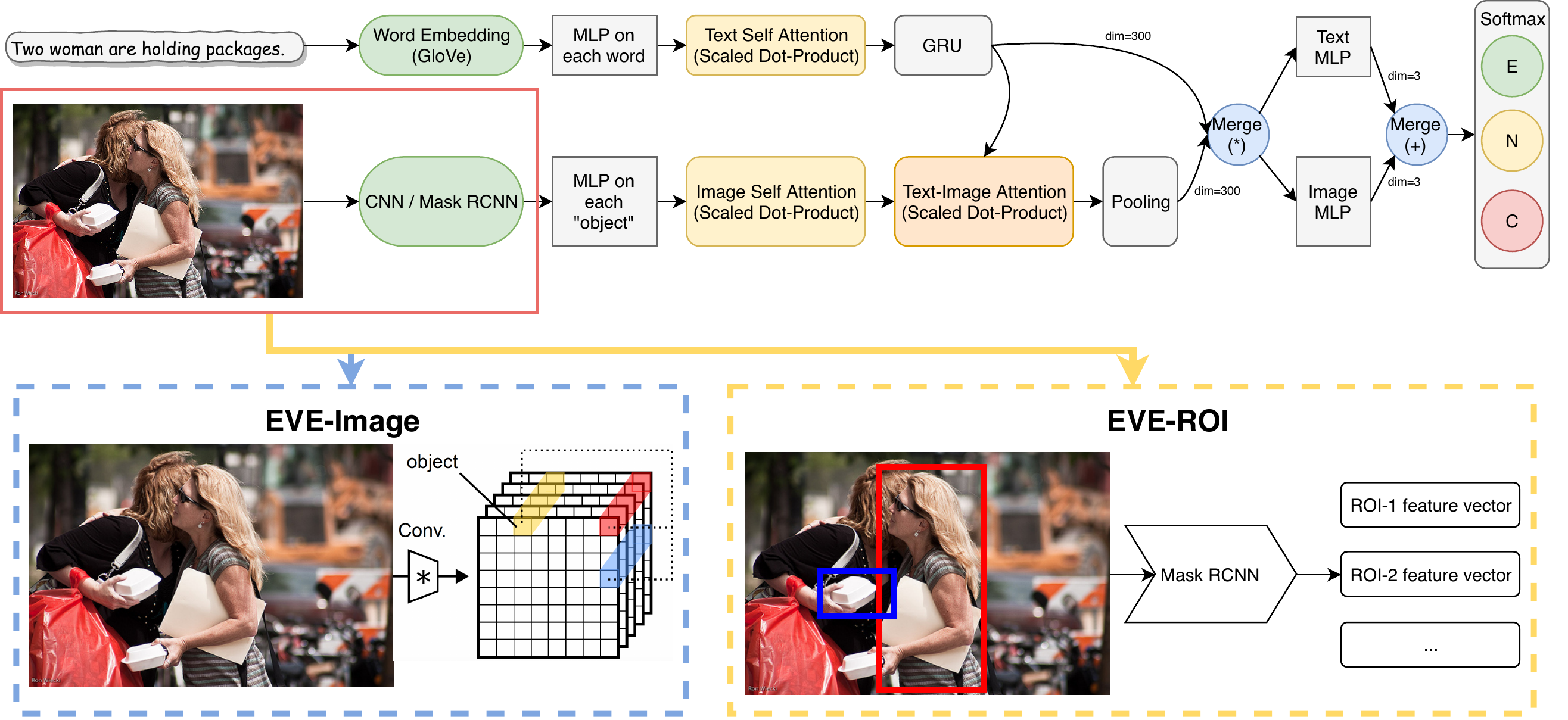}
  \caption{EVE architecture. EVE determines if a hypothesis (text input) is entailed by an image premise (image input). The bottom half shows two methods on image feature extraction, either from the CNN feature maps or object detection ROIs.}
  \label{fig:EVE}
\end{figure}

We develop a new Explainable Visual Entailment architecture (EVE) shown in Figure~\ref{fig:EVE}.
EVE uses the Attention Top-Down/Bottom-Up~\citep{anderson2018bottom} model as a starting point.
The architecture consists of two branches. 
The text branch applies self-attention~\citep{vaswani2017attention} to the word embeddings of a given text hypothesis, then passes the weighted word embedding sequence through gated recurrent units to extract the text features.
Depending on the image feature extraction in the image branch, there are two EVE variants: \textbf{EVE-Image} and \textbf{EVE-ROI}.
The image features captured by EVE-Image come from a pre-trained convolutional neural network (CNN) with $k$ feature maps of dimension $d \times d$. 
The feature vector at each pixel position across the $k$ feature maps 
represents an image region. 
In contrast, EVE-ROI considers regions of interest (ROI) proposals from MASK-RCNN~\citep{he2017mask} to locate prominent objects in images.
The image regions either from EVE-ROI or EVE-Image are also self-attended and further weighted by the text-image attention.
Both the text and image features are finally fused for later prediction.

We apply \textbf{self-attention} to capture the hidden relations between elements in the text and the image feature spaces respectively. 
The intuition of using self-attention is, under a long and complex hypothesis, it is increasingly necessary for the model to be able to attend to only the most relevant words.
The effect of self-attention on the image is similar: image regions that jointly benefit the current prediction receive more attention.
On the other hand, the \textbf{text-image attention} allows the model to select relevant image regions conditioned on the given text hypothesis.

\section{Evaluation on SNLI-VE}
\label{sec:eval}

\begin{table}[h]
\newcommand{\tabincell}[2]{\begin{tabular}{@{}#1@{}}#2\end{tabular}}
\begin{tabular}{rcccccccc}
\toprule
& & \multicolumn{3}{c}{\textbf{Val Acc Per Class (\%)}} & 
\multirow{2}{*}{\tabincell{l}{\textbf{Test} \\ \textbf{Acc (\%)}}} &
\multicolumn{3}{c}{\textbf{Test Acc Per Class (\%)}} \\
\multirow{-2}{*}{\textbf{Model Name}} &
\multirow{-2}{*}{\tabincell{l}{\textbf{Val} \\ \textbf{Acc (\%)}}} & 
\textbf{C}    & \textbf{N} & \textbf{E} & \textbf{} & \textbf{C} & \textbf{N} & \textbf{E} \\
\midrule
\textbf{Hypothesis Only} & 67.04 & 65.45 & 63.36 & 72.31 & 67.01 & 65.85  & 63.78 & 71.40 \\
\textbf{Image Captioning} & 68.14 & 67.3 & 63.12 & 73.99 & 67.47 & 66.75  & 63.56 & 72.07 \\
\textbf{Relational Network} & 67.81 & 68.01 & 63.94 & 71.49 & 68.39 & 69.13 & 65.58 & 70.45 \\
\textbf{Attention Top-Down}  & 70.59  & \textbf{72.94}   & 66.88 & 71.96 & 70.3 & \textbf{72.94}   & 66.63 & 71.34 \\
\textbf{Attention Bottom-Up} & 69.79 & 71.56 & 64.25 & 73.57 & 69.34 & 70.56 & 64.49 & 72.96  \\
\textbf{EVE-Image*} & \textbf{71.40}  & 70.48  & 66.88 & 76.83  & \textbf{71.36}  & 70.61 & 67.17  & \textbf{76.31}  \\
\textbf{EVE-ROI*}  & 71.11 & 66.41 & \textbf{68.2}  & \textbf{78.69}  & 70.21 & 65.63 & \textbf{68.83} & 76.16 \\ 
\bottomrule
\\
\end{tabular}
\caption{Model Performance on SNLI-VE dataset}
\label{table:performance}
\end{table}

We evaluate the performance of EVE against several other baselines over SNLI-VE including the existing state-of-the-art VQA based models. 
Details about the dataset and our experiments are discussed in the supplemental materials. 
The performance results, as listed in Table~\ref{table:performance}, 
involve comparisons between the following models: 

\textbf{Hypothesis Only:} 
This model uses hypotheses only without image premises.
Based on no premises, the model was expected to make random guesses 
but the resulting accuracy is up to 67\%, as reproduced by others~\citep{gururangan2018annotation, vu2018grounded}.
This indicates the performance of our model must exceed the 67\% lower bound to make sense.

\textbf{Image Captioning:} 
Before VE, there are many captioning models~\citep{karpathy2015deep,vinyals2017show,chen2017sca} which can serve as a useful baseline by generating an image caption as the premise and then apply existing TE models for classification.
For this baseline, we use a PyTorch implementation \citep{CaptioningPy} 
which extracts the image features with a pre-trained ResNet152 backbone 
and generates the captions using an LSTM.
The generated text premise is encoded with the input text hypothesis.
Both text features are concatenated for classification.
The model performance achieves a marginally higher accuracy of 68.14\% and 67.47\% on the validation and test sets respectively, implying that the generated image caption premise does not help much. 
After reviewing the generated captions, it is possible that the quality of the generated captions are too poor or missing the necessary information for the TE classifier.
To address this problem, the captioning may be improved by using sophisticated models such as the dense captioning~\citep{johnson2016densecap} but there is no guarantee that every detail in the image potentially described by the hypothesis would be covered.
Nevertheless, the TE classifier could still perform poorly due to the increase in the length of the caption premises.

\textbf{Relational Network:} 
The Relational Network (RN), proposed to tackle the CLEVR dataset considers pairwise feature fusions between different image regions in the CNN feature maps and the question embedding~\citep{santoro2017simple}.
Although RN provides high accuracy on CLEVR, only a marginal improvement is achieved at the accuracy of 67.81\% and 68.39\% on the validation and test splits of SNLI-VE.

\textbf{Attention Top-Down:} 
We also adopt the model from the winner~\citep{anderson2018bottom} of VQA challenge 2017, which applies text-image attention to the image regions in the CNN feature maps based on the question embedding. 
The weighted image features are then projected and fused with the question embedding using dot-product for classification.
This attention based VQA model achieves the best accuracy so far, with 70.59\% and 70.3\% on the validation and test splits, respectively, implying attention can effectively use image premise features.

\textbf{Attention Bottom-Up:} The model design for Attention Bottom-Up is quite similar to Attention Top-Down, except the image features used are the ROIs extracted by a Mask-RCNN ~\citep{he2017mask} implementation \citep{MaskRCNNPy}.
The best performance achieved is 69.79\% and 69.34\% accuracy on the validation and testing splits respectively.
Though we also evaluate the model with more than 10 ROIs, we observe no significant improvement.

\textbf{EVE-Image} and \textbf{EVE-ROI:}
We finally evaluate our model, EVE, as described in Section~\ref{sec:eve}.
EVE-Image achieves the best performance of 71.4\% and 71.36\% accuracy on the validation and test partitions.
EVE-ROI achieves a slightly lower accuracy of 71.11\% and 70.21\% but still better than the counterpart Attention Bottom-Up.
The improvement, even just marginal, is likely attributed to the introduction of self-attention that captures the hidden relations in the same feature space.

\section{Conclusion} This work introduces \emph{visual entailment}, a novel
multimodal task to determine if a text hypothesis is entailed based on the visual information in the image premise.
We build the SNLI-VE dataset providing real-world images from Flickr30k as premises, and the corresponding text hypotheses from SNLI. 
To address VE, we develop EVE and demonstrate its performance over several baselines, including the existing state-of-the-art VQA based models.
The inherent language-bias induced by SNLI~\citep{gururangan2018annotation} serves as a strong baseline.
The SNLI-VE dataset is publicly available at \url{https://github.com/necla-ml/SNLI-VE}.

\clearpage

\section*{Acknowledgments}
Ning Xie and Derek Doran were supported by the Ohio Federal Research Network project \textit{Human-Centered Big Data}. 
Any opinions, findings, and conclusions or recommendations expressed in this article are those of the author(s) and do not necessarily reflect the views of the Ohio Federal Research Network.

{\small
\bibliographystyle{abbrvnat}
\bibliography{egbib}
}


\section*{Supplementary Materials}

\paragraph{Dataset statistics.}
    The original SNLI dataset split does not consider the arrangement of the original caption images.
Therefore, the same image may appear in both training and test sets if directly adapted to VE.
To address the issue, we disjointedly partition SNLI-VE by images following the splits in \citep{gong2014improving} and make sure that each class instances are balanced across the training, validation, and test sets as shown in Table~\ref{table:dataset}.

\begin{table*}[h]
\centering
\begin{tabular}{rrrrcccccc}
\toprule
\textbf{} & \textbf{Training} & \textbf{Validation} & \textbf{Testing} \\
\midrule
\textbf{\#Images} & 29,783 & 1,000 & 1,000 \\
\textbf{\#Entailment} & 176,932 & 5,959 & 5,973 \\
\textbf{\#Neutral} & 176,045 & 5,960 & 5,964 \\
\textbf{\#Contradiction} & 176,550 & 5,939 & 5,964 \\
\textbf{Vocabulary Size} & 29,550 & 6,576 & 6,592 \\
\bottomrule
\end{tabular}
\caption{SNLI-VE statistics: number of images, per class examples and vocabulary size by split.}
\label{table:dataset}
\end{table*}

\paragraph{Implementation details.}
The proposed EVE model is implemented in PyTorch.
We use the pre-trained GloVe.6B.300D~\citep{pennington2014glove} for word embedding, where 6B is the corpus size and 300D is the embedding dimension.
The image features used for EVE-Image are generated from a pre-trained ResNet101.
The ROI features used for EVE-ROI are extracted using the Mask-RCNN implementation~\citep{MaskRCNNPy}.
The Adam optimizer is used for training with a batch size of 64.
Adaptive learning rate is applied with both initial value and weight decay set to be 0.0001.

\end{document}